\documentclass{article}

\PassOptionsToPackage{square, numbers}{natbib}
\usepackage{graphicx}
\usepackage{float}
\usepackage{gensymb}
\usepackage{amsmath}
\usepackage[preprint]{neurips_2022}

\usepackage[utf8]{inputenc} 
\usepackage[T1]{fontenc}    
\usepackage{hyperref}       
\usepackage{url}            
\usepackage{booktabs}       
\usepackage{amsfonts}       
\usepackage{nicefrac}       
\usepackage{microtype}      
\usepackage{xcolor}         

\title{\textsc{DRLComplex}: Reconstruction of protein quaternary structures using deep reinforcement learning}

\author{%
  Elham Soltanikazemi\thanks{Equal Contribution}\ , Raj S. Roy\thanks{Equal Contribution}\ , Farhan Quadir, Nabin Giri, Alex Morehead, Jianlin Cheng \\
  Department of Electrical Engineering \& Computer Science\\
  University of Missouri\\
  Columbia, MO 65211 \\
  \texttt{\{esdft, rsr3gt, fqg7h, ngzvh, acmwhb, chengji\}@missouri.edu} \\
}

\begin{document}

\maketitle

\begin{abstract}
Predicted inter-chain residue-residue contacts can be used to build the quaternary structure of protein complexes from scratch. However, only a small number of methods have been developed to reconstruct protein quaternary structures using predicted inter-chain contacts. Here, we present an agent-based self-learning method based on deep reinforcement learning (\textsc{DRLComplex}) to build protein complex structures using inter-chain contacts as distance constraints. We rigorously tested \textsc{DRLComplex} on two standard datasets of homodimeric and heterodimeric protein complexes (i.e., the CASP-CAPRI homodimer and Std\_32 heterodimer datasets) using both true and predicted interchain contacts as inputs. Utilizing true contacts as input, \textsc{DRLComplex} achieved high average TM-scores of 0.9895 and 0.9881 and a low average interface RMSD (I\_RMSD) of 0.2197 and 0.92 on the two datasets, respectively. When predicted contacts are used, the method achieves TM-scores of 0.73 and 0.76 for homodimers and heterodimers, respectively. Our experiments find that the accuracy of reconstructed quaternary structures depends on the accuracy of the contact predictions. Compared to other optimization methods for reconstructing quaternary structures from inter-chain contacts, \textsc{DRLComplex} performs similar to an advanced gradient descent method and better than a Markov Chain Monte Carlo simulation method and a simulated annealing-based method, validating the effectiveness of \textsc{DRLComplex} for quaternary reconstruction of protein complexes.
\end{abstract}

\section{Introduction}
In cells, a protein chain folds into a three-dimensional (3D) shape referred to as a tertiary structure. The tertiary structures of two or more protein chains often interact to form a protein complex, a conformation often referred to as a protein quaternary structure. As the function of a single protein chain or protein complex is largely determined by its 3D structure, predicting protein structures from their corresponding amino acid sequences is important for studying the function of proteins. However, the determination of protein quaternary structures using experimental techniques such as X-ray crystallography remains time-consuming and expensive and, as such, can only be applied to a small portion of proteins of interest. Therefore, accurate computational prediction of protein structures from sequences has become a holy grail of computational biochemistry, drug design, biophysics, and bioinformatics over the last few decades.\par

Recently, deep learning (DL) has driven substantial progress in protein tertiary structure prediction in the recent years \citep{jones2018high, adhikari2018dncon2, li2019respre, senior2019protein, yang2020improved, wu2021deepdist}, which has been demonstrated in the biennial Critical Assessment of Protein Structure Prediction (CASP) competition \citep{kryshtafovych2014casp, kryshtafovych2019critical, won2019assessment}. A major breakthrough was made in the $14$th round of CASP by AlphaFold 2 \citep{jumper2020alphafold}, a method that can predict tertiary structures of many proteins with high accuracy. Despite the notable progress in tertiary structure prediction, the prediction of protein quaternary structures (i.e., protein structures consisting of two or more protein chains) is still in early stages of development \citep{zeng2018complexcontact, hou2020multicom, quadir2021deepcomplex, quadir2021dncon2_inter, yan2021accurate, roy2022deep, xie2022deep}. Towards this end, we propose \textsc{DRLComplex}, a novel deep reinforcement learning (DRL) approach to reconstructing quaternary structures of protein complexes. In particular, \textsc{DRLComplex} generates quaternary structures of protein dimers using predicted inter-chain contacts and known (or predicted) tertiary structures as input to a self-play mechanism for geometric reconstruction, making it the \textbf{first of its kind} in DL-based complex structure modeling.\par

\section{Related Work}
We now proceed to describe works relevant to DL-based complex structure modeling and prediction.

\textbf{Quaternary structure prediction taxonomy.} Existing methods for quaternary structure prediction can currently be subdivided into two categories: \textit{ab initio} methods \citep{lyskov2008rosettadock, pierce2014zdock, quadir2021deepcomplex, park2021galaxyheteromer, evans2021protein, soltanikazemi2022distance} and template-based methods \citep{tuncbag2012fast, guerler2013mapping}. Template-based approaches typically search protein structure databases for similar templates for a protein target and predict its quaternary structure based on the available template structures. While such approaches work well if suitable structural templates are available, it does not work well for a majority of protein complexes, as template structures often are not available for such types of proteins. As such, we restrict the scope of our review of related works to \textit{ab initio} methods.\par

\textbf{\textit{Ab initio} quaternary structure reconstruction methods.} Different \textit{ab initio} methods based on the energy optimization and machine learning prediction of inter-chain contacts have been developed and tested in the Critical Assessment of Protein Interaction (CAPRI) experiments \citep{lensink2021prediction}. For example, RosettaDock is a popular docking tool that leverages Markov Chain Monte-Carlo (MC) based algorithms for energy minimization to dock protein chains \citep{lyskov2008rosettadock}. Likewise, Zdock \citep{pierce2014zdock} uses a combination of various potential scoring criteria to select quaternary structure models generated by a Fourier transformation method. Recently, a gradient descent structure optimization method, GD \citep{soltanikazemi2022distance}, and simulated annealing method, ConComplex \citep{quadir2021deepcomplex}, have been developed to build quaternary structures from predicted inter-chain residue-residue contacts and tertiary structures of protein chains. Due to recent development of accurate deep learning methods for inter-chain contact prediction such as DeepInteract \cite{morehead2021geometric}, GLINTER \citep{xie2022deep}, and DRCon \citep{roy2022deep}, the predictions of such methods have enabled existing quaternary structure predictors to generate high-quality quaternary structures for certain proteins.\par

\textbf{\textit{Ab initio} quaternary structure prediction methods.} The latest advancements in the field of quaternary structure prediction, however, are extensions of end-to-end deep learning methods for tertiary structure prediction \citep{baek2021accurate, jumper2020alphafold}. For instance, AlphaFold-Multimer \citet{evans2021protein} adapts the deep learning architecture of AlphaFold 2 to take as input concatenated multiple sequence alignments of protein monomers found within protein multimers as well as available template structures to predict corresponding quaternary structures. This method outperforms existing docking methods according to two benchmark studies. However, AlphaFold-Multimer is currently only able to predict accurate quaternary structures for a portion of protein complexes.\par

\textbf{Reinforcement learning methods.} In parallel to many advances in computational biology, advancements in DRL have been demonstrated by methods such as AlphaGo \citep{silver2016mastering} for playing GO games, defeating the GO world champion multiple times. DRL has also achieved similar success in playing other games such as Atari games \citep{mnih2013playing} as well as StarCraft II \citep{vinyals2019grandmaster}. Though reinforcement learning (RL) has commonly been used in the context of gaming and robotics, it has rarely been used in the field of computational biology \citep{wang2018deep, hou2019using, bocicor2011reinforcement} and, until now, has not been used for protein complex structure modeling.\par

\textbf{Contributions.}\\ Our work builds upon that of previous works by making the following contributions.

\begin{itemize}
\item We provide the \textit{first} example of applying deep reinforcement learning to the task of protein complex structure modeling.
\item We provide the \textit{first} example of applying a self-play reinforcement learning mechanism within the field of computational biology.
\item We introduce \textsc{DRLComplex}, showcasing its state-of-the-art performance in reconstructing protein quaternary structures.
\end{itemize}

\section{\textsc{DRLComplex} Model}
\begin{figure}[!t]
  \centering
  \includegraphics[width=0.7\linewidth]{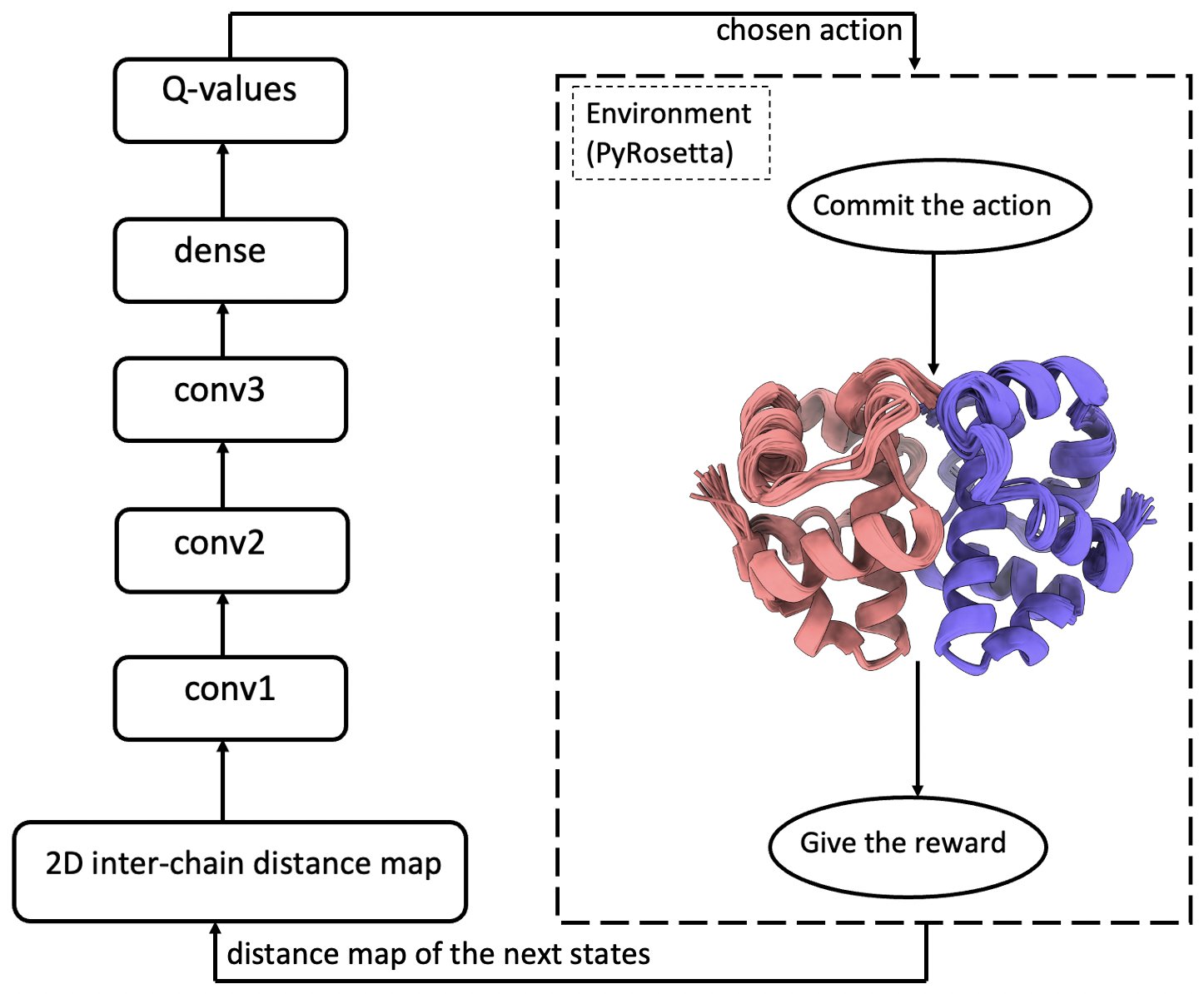}
  \caption{The deep network to predict the values of six actions given the inter-chain distance map ($S$) of the state of a dimer as input. The first convolution layer has 16 filters of size 3 $\times$ 3 with stride 2. The second layer contains 32 filters with stride 2, and the filter size is 5 $\times$ 5. The third convolutional layer has 64 3 $\times$ 3 filters with stride 2. The rectified linear unit (ReLU) function is used by all the three convolutional layers. The output of the third convolutional layer is then fed into a fully connected layer to generate hidden features, which are used by the fully connected output layer with the linear activation function to predict Q-values of the six actions. }
  \label{figure5}
\end{figure}

We now turn to describe our new \textsc{DRLComplex} model, as illustrated in Figure \ref{figure5}.

\textbf{Problem formulation.} \textsc{DRLComplex} uses deep reinforcement learning techniques to automatically construct the quaternary structure of protein dimers by adjusting the position of one protein chain (ligand) with respect to another one held fixed (receptor) to produce structural conformations similar to the native (true) structure. Specifically, an artificial intelligence (AI) \textit{agent} is trained to choose \textit{actions} according to both \textit{intermediate reward} and \textit{long-term reward} to modify the \textit{state} (structure) of a protein dimer in a modeling \textit{environment}. The state ($S$) of a protein dimer is represented by two equivalent forms: the 3D coordinates of atoms and their inter-chain distance map. The former is suitable for generating new structures in the environment by applying actions to such coordinates, and the latter is suitable for a deep learning model to predict the values of possible actions. Here, we represent inter-chain distance maps as a two-dimensional (2D) matrix ($M$) with a size of $L_{1}$ $\times$ $L_{2}$, where $L_{1}$ is the sequence length of the ligand and $L_{2}$ is the sequence length of the receptor. The cells of such a matrix contain the distances between residues' $C_{\beta}$ atoms within a dimer's ligand and receptor structures, respectively.\par

\textbf{Network design.} In \textsc{DRLComplex}, we designed six actions for our agent, including three translations with a step size of 1 {\AA} and three rotations with a step size of 1$\degree$ along the x, y, and z-axis (i.e., forward and backward motions) to adjust the position of one chain (e.g., the ligand) against the other one (e.g., the receptor). We then use a deep convolutional neural network (i.e., Q) to approximate the Q-function of our RL context to predict Q-values (and estimate immediate and future rewards) of our six possible actions, as shown in Figure \ref{figure5} for an input state $S$. We refer to this deep network as an agent that predicts the value of the actions, values which are then used to choose the next action to change the state of a protein dimer structure.\par

\begin{figure}[!t]
  \centering
  \includegraphics[width=0.7\linewidth]{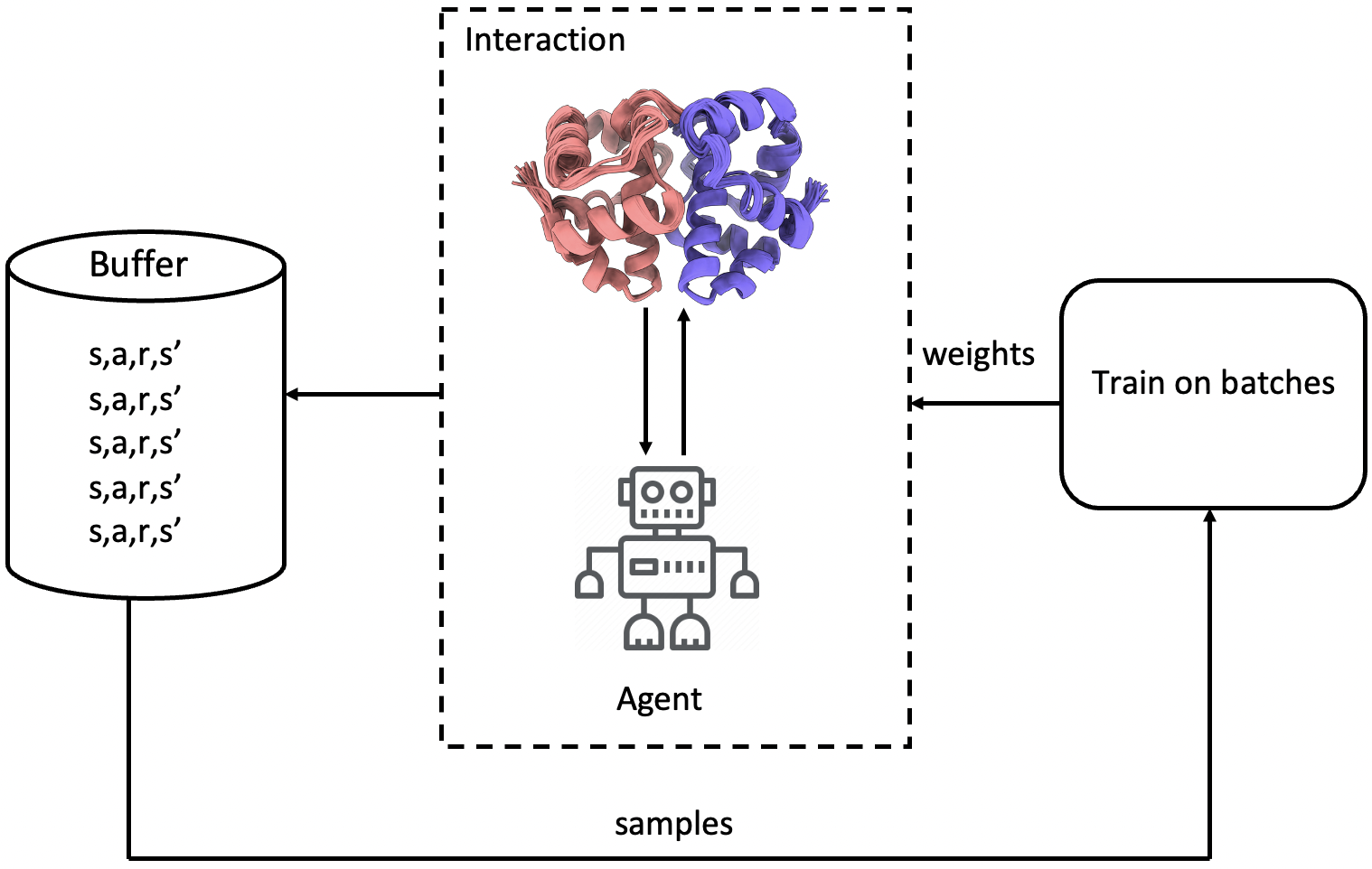}
  \caption{Training of the deep network with Experience Replay for a protein dimer. The agent interacts with the environment to generate a sequence of experiences (i.e., <$s,a,r,s^{'}$> vectors). These experiences are then stored in a buffer. If the buffer is completely filled, some of the earliest experiences are deleted to create space for new experiences. Here, $K$ transition experiences are sampled from the buffer to train the deep network. The process is repeated until the system converges (i.e., when a final state is reached).}
  \label{figure6}
\end{figure}

The deep convolutional neural network is trained using a sequence of examples (state ($S$), action ($A$), reward ($R$), next State ($S^{'}$)), which are automatically obtained by the agent continuously interacting with the modeling environment (implemented within PyRosetta \citep{chaudhury2010pyrosetta}) to choose actions to adjust the position of the ligand according to an $\varepsilon$-\textit{greedy} policy. Given a state $S$ at each step, the agent selects either an action having the highest value predicted by the deep network with a probability of 1 - $\varepsilon$ or a random action with $\varepsilon$ probability. The action ($A$) is then applied to the 3D structure of the current state ($S$) by the environment ($E$) to generate a new 3D structure corresponding to the next state ($S^{'}$). The immediate reward ($R$) of the action is calculated as the difference between the quality of ($S^{'}$) and ($S$) with respect to the target state $S^{*}$, which can be either the true structure of the dimer or the predicted inter-chain contact map provided to the agent at the beginning. The process is repeated to generate many examples. Notably, we use Experience Replay to train the deep network, which stores the agent's experience at each step ($S, A, R, S^{'}$) within a computational buffer. Training with Experience Replay is illustrated in Figure \ref{figure6}.\par

\begin{figure}[!t]
  \centering
  \includegraphics[width=0.5\linewidth]{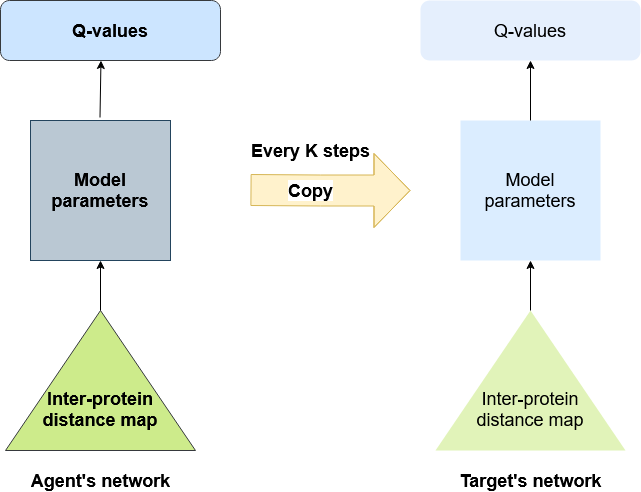}
  \caption{The agent's deep network (Q) and the target network (Q$_{reference}$) to generate reference Q-values of actions to train Q.}
  \label{figure7}
\end{figure}

We used the immediate reward and the output of a \textit{target network} (Q$_{target}$) to generate the reference Q-value (Q$_{reference}$(S, a)) for an action $a$ and a state $S$ to serve as the labels to train the deep network (Q) of the agent. The target network uses a copy of the weights of the deep network of the agent at every $k$ steps to predict the Q-value of each action $a{'}$ (Q$_{target}$($s^{'}, a^{'}$)) for the next state $S^{'}$ (shown in Figure \ref{figure7}), a Q-value that is then used to estimate the future value of each action. The reference Q-value for $S$ and $a$ (Q$_{reference}$(s, a)) is the immediate reward $r(S,a)$ plus the predicted-highest future value, including a multiplicative discount factor $\gamma$ (i.e, Q$_{reference}$(s, a) = r(s, a) + $\gamma \times max_{a^{'}} Q_{target}(s^{'}, a^{'}$). The maximum future value is the maximum Q-value predicted by the target network for the next state $S^{'}$ and any possible action $a^{'}$.\par

\textbf{State evaluation strategies.} We tested two different strategies to evaluate the quality of a state (S) to calculate the immediate reward: (1) the root mean square distance (RMSD) between the 3D structure of S and the true structure (S$^{*}$), and (2) the contact energy - the agreement between the distance map of S and the inter-chain contact map provided as the initial input. We note that only the latter is used in our final experiments, as the true structure is typically unknown in many real applications. For our first strategy, the RMSD of the current structure of S and the next state (S$^{'}$) is calculated as RMSD$_{s}$ and RMSD$_{s^{'}}$ and the immediate reward $r(S, a)$ is equal to RMSD$_{s}$ - RMSD$_{s^{'}}$. If the quality of the next state generated by action $a$ is better (i.e., RMSD$_{s^{'}}$ < RMSD$_{s}$), the agent receives a positive reward; otherwise, it gets a negative reward (i.e., a penalty).\par

For our second strategy, the immediate reward is the contact energy of $S$ minus the contact energy of S$^{'}$. The contact energy function measuring the satisfaction of a contact between any two residues is defined as shown in Equation \ref{eq2}.\par

\begin{equation}
  f(x) =
    \begin{cases}
      (\frac{x-lb}{sd})^{2} & x < lb\\
      0 & lb \le x \le ub\\
      (\frac{x-ub}{sd})^{2} & ub < x < x + sd \\
      \frac{1}{sd}(x - (ub + sd)) & x> ub + sd
    \end{cases}   
    \label{eq2}
\end{equation}

In Equation \ref{eq2}, $lb$ and $ub$ represent the lower bound and upper bound of the distance ($x$) between two residues that are predicted to be in contact. Two residues are considered in contact if the distance between their closet heavy atoms is less than or equal to 6 {\AA}. $sd$ is the standard deviation, which is set to 0.1. Based on this cost function, if the distance between two residues predicted to be in contact is $\le$ 6 {\AA}, i.e., the contact restraint is satisfied, and the contact energy is 0. Otherwise, it is a positive value. The complete contact energy function for the structural model of a state (S or S$^{'}$) is the sum of the energies for all predicted inter-chain contacts used in modeling (called contact energy). The immediate reward is the contact energy of (S) minus that of S$^{'}$. The goal is to find actions to increase the satisfaction of contacts in the structure (i.e, to reduce the contact energy).\par

\begin{figure}[H]
  \centering
  \includegraphics[width=1\linewidth]{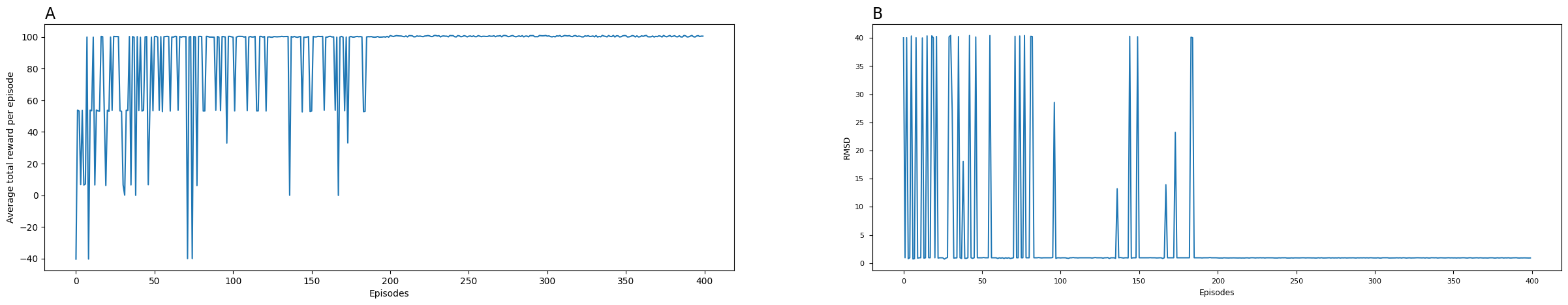}
  \caption{The total rewards and the RMSD of the reconstructed structures at each episode for a protein dimer (PDB Code: 1A2D) using the contact energy to calculate the reward. (A) The average total reward at each episode averaged over several self-learning runs (games). (B) The RMSD of the predicted structure at the end of each episode.}
  \label{figure8}
\end{figure}

\textbf{State evaluation experiment.} As an example to illustrate the effect of our state evaluation strategies on \textsc{DRLComplex}'s reconstruction performance, Figure \ref{figure8} shows how the accumulated award and the RMSD of the reconstructed structure for a protein dimer (PDB Code: 1A2D) change across episodes in the self-learning process by using the contact energy to calculate the reward. Notably, the accumulated reward converges in the last 200 episodes. The self-learning mechanism (i.e., game) finishes by generating a structure with an RMSD of 0.94 {\AA}. This experiment demonstrates that a high-quality structure can be generated using DRLComplex to maximize rewards based on predicted inter-chain contacts. This deep network was trained for 100K steps, using a batch size of 32. Furthermore, the target network was updated every 500 steps. It is worth noting that the deep network is trained specifically for each target through self-learning. Therefore, the inference time for each target involves the training time as well.\par

\section{Experiments}
\subsection{Evaluation Setup}
\textbf{Test Datasets.} We conduct all subsequent experiments for model evaluation on two standard protein dimer datasets: CASP\_CAPRI and Std\_32. The CASP\_CAPRI dataset consists of 28 homodimer \citep{yan2021accurate} targets, and the Std\_32 contains 32 heterodimer targets \citep{soltanikazemi2022distance}. We note that we exclude one of the Std\_32 targets from our experiments, as this target (PDB Code: 1IXRA\_1IXRC) contains no inter-chain contacts between its ligand and receptor structures.\par

\textbf{Baselines.} We compare \textsc{DRLComplex} with four existing methods, namely gradient descent (GD) \citep{soltanikazemi2022distance}, Markov Chain Monte Carlo simulation (MC), EquiDock \citep{ganea2021independent}, and CNS \citep{brunger1998crystallography, brunger2007version}, the simulated annealing simulation used in ConComplex \citep{quadir2021deepcomplex} that constructs quaternary structures. Here, we note that EquiDock is an equivariant neural network that directly predicts quaternary structures of dimers without using inter-chain contacts as input. In contrast, \textsc{DRLComplex}, GD, MC, and CNS all reconstruct quaternary structures from inter-chain contacts.\par

\textbf{Evaluation Methods.} Each method is evaluated under three scenarios: the optimal scenario, the suboptimal scenario, and the realistic scenario. In the optimal scenario, true inter-chain contacts are extracted from the native quaternary structures of protein dimers and used as input to guide the assembly of the true tertiary structures of monomers in their bound state into quaternary structures. Here, an inter-chain contact is a pair of residues from the two chains in a dimer in which the shortest distance between their heavy atoms is less than or equal to 6 {\AA} \citep{quadir2021dncon2_inter}. In the suboptimal scenario, the predicted inter-chain contacts together with the true tertiary structure of monomers in their bound state are used as input. We used two inter-chain contact predictors for this scenario, DRCon (6 {\AA} version) to predict inter-chain contacts for the CASP\_CAPRI homodimer dataset and GLINTER \citep{xie2022deep} to predict inter-chain contacts for the Std\_32 heterodimer dataset. In the final and most realistic scenario, both the predicted inter-chain contacts as well as the unbound tertiary structures of monomers predicted by AlphaFold 2 \citep{jumper2020alphafold} are used as input.\par

The performance of these methods is evaluated by comparing their reconstructed structures with native/true structures of the dimers using TMalign \citep{zhang2005tm} and DockQ \citep{basu2016dockq}. The evaluation metrics include TM-score, root-mean-square deviation (RMSD) of the entire structure with respect to the native structure, the fraction of the native inter-chain contacts in the reconstructed structures ($f_{nat}$, interface RMSD (I\_RMSD), and Ligand(L\_RMSD).\par

\textbf{Implementation Details.} As shown in Equation \ref{eq1}, we used the average squared error as the loss $\mathcal{L}$ to train \textsc{DRLComplex}'s Q network, where $N$ is the number of states. We also used a stochastic gradient descent optimizer with a mini-batch size of 64 for training. During each experiment, $\varepsilon$ of the $\varepsilon$-policy is decreased from 1 to 0.1 for the first 100,000 iterations via exponential decay (decay factor = 0.99) and is fixed at 0.1 thereafter. We trained \textsc{DRLComplex} using three Nvidia GPUs, each with 32 GB, on a system with 350GB of RAM in total.\par

\begin{equation}
    \mathcal{L} = \frac{1}{N}\Sigma_{S_{i}} (Q(s,a) - Q_{reference}(s_{i}, a_{i}))^{2}
    \label{eq1}
\end{equation}

\subsection{Results}
\textbf{Evaluation on the Std\_32 Heterodimer Dataset.}
\begin{table}[H]
  \caption{Average mean of RMSD, TM-score, $f_{nat}$, I\_RMSD, and L\_RMSD results of \textsc{DRLComplex}, GD, MC, and CNS on the Std\_32 dataset with true inter-chain contacts and true tertiary structure of monomers as inputs. Bold numbers denote the best results.}
  \label{table1}
  \centering
  \begin{tabular}{llllll}
    \toprule
    \multicolumn{6}{c}{Optimal Scenario}                   \\
    \cmidrule(r){2-6}
    \textbf{Methods} & \textbf{TM-score} ( $\uparrow$ ) & \textbf{RMSD} ( $\downarrow$ ) & \textbf{f$_{nat}$} (\%, $\uparrow$ ) & \textbf{I\_RMSD} ( $\downarrow$ ) & \textbf{L\_RMSD} ( $\downarrow$ ) \\
    \midrule
    \textsc{DRLComplex} & \textbf{0.98}  & \textbf{0.88} & 90.03 & \textbf{0.92} & \textbf{2.15}\\
    GD & 0.95 & 2.9 & \textbf{92.43} & 1.99 & 7.16 \\
    MC & 0.94 & 3.1 & 92.24 & 2.2 & 7.18 \\
    CNS & 0.82 & 10.04 & 69.13 & 3.71 & 14.99 \\
    \bottomrule
  \end{tabular}
\end{table}

\begin{table}[H]
  \caption{Average mean RMSD, TM-score, $f_{nat}$, I\_RMSD, and L\_RMSD of the models reconstructed by \textsc{DRLComplex}, GD, MC, CNS and EquiDock on Std\_32 with predicted inter-chain contacts and true tertiary structure of monomers as inputs. Bold numbers denote the best results.}
  \label{table2}
  \centering
  \begin{tabular}{llllll}
    \toprule
    \multicolumn{6}{c}{Suboptimal Scenario}                   \\
    \cmidrule(r){2-6}
    \textbf{Methods} & \textbf{TM-score} ( $\uparrow$ ) & \textbf{RMSD} ( $\downarrow$ ) & \textbf{f$_{nat}$} (\%, $\uparrow$ ) & \textbf{I\_RMSD} ( $\downarrow$ ) & \textbf{L\_RMSD} ( $\downarrow$ ) \\
    \midrule
    \textsc{DRLComplex} & \textbf{0.76}  & 13.93 & \textbf{19.64} & \textbf{13.68} & \textbf{34.16}\\
    GD & 0.75 & \textbf{13.92} & 16.68 & 13.72 & 34.17 \\
    MC & 0.73 & 14.14 & 13.88 & 13.82 & 35.36 \\
    CNS & 0.68 & 17.09 & 17.26 & 15.81 & 43.12 \\
    EquiDock & 0.61 & 18.53 & 4.95 & 14.98 & 36.11 \\
    \bottomrule
  \end{tabular}
\end{table}

\begin{table}[H]
  \caption{Average mean RMSD, TM-score, $f_{nat}$, I\_RMSD, and L\_RMSD of the models reconstructed by the \textsc{DRLComplex}, GD, MC, CNS and EquiDock methods on Std\_32 with predicted inter-chain contacts and predicted tertiary structures of monomers as inputs. Bold numbers denote the best results.}
  \label{table3}
  \centering
  \begin{tabular}{llllll}
    \toprule
    \multicolumn{6}{c}{Realistic Scenario}                   \\
    \cmidrule(r){2-6}
    \textbf{Methods} & \textbf{TM-score} ( $\uparrow$ ) & \textbf{RMSD} ( $\downarrow$ ) & \textbf{f$_{nat}$} (\%, $\uparrow$ ) & \textbf{I\_RMSD} ( $\downarrow$ ) & \textbf{L\_RMSD} ( $\downarrow$ ) \\
    \midrule
    \textsc{DRLComplex} & \textbf{0.74}  & \textbf{14.53} & \textbf{11.80} & 13.86 & 36.18\\
    GD & 0.73 & 14.54 & 11.71 & \textbf{13.85} & \textbf{36.17} \\
    MC & 0.70 & 15.06 & 11.25 & 14.60 & 36.66 \\
    CNS & 0.66 & 19.02 & 3.78 & 14.74 & 37.91 \\
    EquiDock & 0.59 & 18.63 & 3.55 & 14.42 & 35.97 \\
    \bottomrule
  \end{tabular}
\end{table}

The performance of all the methods on the Std\_32 heterodimer dataset is reported in Table \ref{table1}, Table \ref{table2}, and Table \ref{table3} for optimal, suboptimal, and realistic scenarios, respectively. In the optimal scenario, Table \ref{table1} shows that \textsc{DRLComplex} outperforms all other the methods in all the evaluation metrics except for $f_{nat}$. In particular, the difference in terms of RMSD is even more pronounced. For example, the average RMSDs of \textsc{DRLComplex}, GD, MC, and CNS are 0.88 {\AA}, 2.9 {\AA}, 3.1 {\AA}, and 10.04 {\AA}, respectively.\par

In the suboptimal scenario, Table \ref{table2} shows that \textsc{DRLComplex}'s performance is better overall than all other methods. For instance, the TM-score of the \textsc{DRLComplex} is 1.33\%, 4.1\%, 11.7\%, and 24.5\% better than that of GD, MC, CNS, and EquiDock, respectively.\par

Similarly, regarding the realistic scenario, Table \ref{table3} demonstrates that \textsc{DRLComplex}'s performance is better overall than other methods, as it has the best performance in terms of TM-score, RMSD, and $f_{nat}$. As shown in Table \ref{table6}, the TM-score of \textsc{DRLComplex} is 1.3\%, 5.7\%, 12.2\% and 25.4\% better than that of GD, MC, CNS, and EquiDock, respectively.\par

\textbf{Performance on CASP\_CAPRI Homodimer Dataset.}
\begin{table}[H]
  \caption{Mean RMSD, TM-score, $f_{nat}$, I\_RMSD, and L\_RMSD of the \textsc{DRLComplex}, GD, MC, and CNS on 28 homodimers in CASP\_CAPRI datase using true inter-chain contacts and true tertiary structures as inputs. Bold numbers denote the best results.}
  \label{table4}
  \centering
  \begin{tabular}{llllll}
    \toprule
    \multicolumn{6}{c}{Optimal Scenario}\\
    \cmidrule(r){2-6}
    \textbf{Methods} & \textbf{TM-score} ( $\uparrow$ ) & \textbf{RMSD} ( $\downarrow$ ) & \textbf{f$_{nat}$} (\%, $\uparrow$ ) & \textbf{I\_RMSD} ( $\downarrow$ ) & \textbf{L\_RMSD} ( $\downarrow$ ) \\
    \midrule
    \textsc{DRLComplex} & \textbf{0.9895}  & \textbf{0.3753} & \textbf{99.05} & \textbf{0.2197} & \textbf{0.8235}\\
    GD & \textbf{0.9895} & \textbf{0.3753} & 99.03 & 0.3468 & \textbf{0.8235}\\
    MC & 0.9631 & 1.2089 & 78.91 & 1.1611 & 2.8897 \\
    CNS & 0.9234 & 2.003 & 73.45 & 3.9234 & 4.6841 \\
    \bottomrule
  \end{tabular}
\end{table}

\begin{table}[H]
  \caption{Mean RMSD, TM-score, $f_{nat}$, I\_RMSD, and L\_RMSD of the \textsc{DRLComplex}, GD, MC, CNS, and EquiDock on 28 homodimers in CASP\_CAPRI dataset using predicted inter-chain contacts and true tertiary structures of monomers as inputs. Bold numbers denote the best results.}
  \label{table5}
  \centering
  \begin{tabular}{llllll}
    \toprule
    \multicolumn{6}{c}{Suboptimal Scenario}                   \\
    \cmidrule(r){2-6}
    \textbf{Methods} & \textbf{TM-score} ( $\uparrow$ ) & \textbf{RMSD} ( $\downarrow$ ) & \textbf{f$_{nat}$} (\%, $\uparrow$ ) & \textbf{I\_RMSD} ( $\downarrow$ ) & \textbf{L\_RMSD} ( $\downarrow$ ) \\
    \midrule
    \textsc{DRLComplex} & 0.73  & 11.88 & 32.21 & 10.43 & 26.52\\
    GD & \textbf{0.74} & \textbf{11.09} & \textbf{35.51} & \textbf{9.66} & \textbf{25.21}\\
    MC & 0.72 & 12.04 & 32.65 & 10.47 & 26.98 \\
    CNS & 0.62 & 14.55 & 28.32 & 14.37 & 36.25 \\
    EquiDock & 0.56 & 18.57 & 6.51 & 14.5 & 35.24 \\
    \bottomrule
  \end{tabular}
\end{table}

\begin{table}[H]
  \caption{Mean RMSD, TM-score, $f_{nat}$, I\_RMSD, and L\_RMSD of the \textsc{DRLComplex}, GD, MC, CNS, and EquiDock  on 28 homodimers in CASP\_CAPRI dataset using predicted inter-chain contacts and predicted tertiary structures of monomers as inputs. Bold numbers denote the best results.}
  \label{table6}
  \centering
  \begin{tabular}{llllll}
    \toprule
    \multicolumn{6}{c}{Realistic Scenario}                   \\
    \cmidrule(r){2-6}
    \textbf{Methods} & \textbf{TM-score} ( $\uparrow$ ) & \textbf{RMSD} ( $\downarrow$ ) & \textbf{f$_{nat}$} (\%, $\uparrow$ ) & \textbf{I\_RMSD} ( $\downarrow$ ) & \textbf{L\_RMSD} ( $\downarrow$ ) \\
    \midrule
    \textsc{DRLComplex} & 0.64 & 12.18 & 27.1 & 10.73 & 26.54\\
    GD & \textbf{0.69} & \textbf{12.15} & \textbf{28.05} & \textbf{10.69} & \textbf{26.50}\\
    MC & 0.63 & 12.78 & 26.81 & 11.89 & 28.92 \\
    CNS & 0.62 & 14.55 & 13.58 & 12.62 & 31.69 \\
    EquiDock & 0.50 & 26.59 & 22.27 & 18.39 & 44.66 \\
    \bottomrule
  \end{tabular}
\end{table}

The average performance of methods on the CASP\_CAPRI homodimer dataset for the optimal, suboptimal, and realistic scenarios are reported in Tables \ref{table4}, \ref{table5}, and \ref{table6}, respectively. In the optimal scenario, Table \ref{table4} shows that \textsc{DRLComplex} performs equally well or better than all other predictors in terms of all evaluation metrics, followed by GD. For 10 of the 28 targets, \textsc{DRLComplex} reconstructs quaternary structures with a perfect $f_{nat}$ score of 1.0. Here, \textsc{DRLComplex}'s average fraction of true inter-chain contacts recalled in the reconstructed structures ($f_{nat}$) is 99.05 \%, indicating that \textsc{DRLComplex} reconstructs highly accurate quaternary structures for dimers when an optimal input is provided. We note that EquiDock was not included in this scenario as it does not use any inter-chain contacts as input but rather directly predicts a translation and rotation of the ligand with respect to the receptor.\par

In the sub-optimal scenario, Table \ref{table5} shows that GD, \textsc{DRLComplex}, and MC's performance is close, while they all outperform CNS and EquiDock by a large margin. For instance, the average TM-score of \textsc{DRLComplex} is 17.7\% and 30.4\% higher than that of CNS and EquiDock, respectively.\par

Lastly, in the realistic scenario, Table \ref{table6} shows that GD performs best here, with \textsc{DRLComplex} following as the second-best method. Nonetheless, here \textsc{DRLComplex} outperforms all other methods. For example, the TM-score of the complexes constructed by \textsc{DRLComplex} is 1.59\%, 3.22\%, and 28\% better than that of MC, CNS, and EquiDock, respectively. We note that here the TM-score of \textsc{DRLComplex} drops from 0.989 to 0.73 since in this case, its predicted inter-chain contacts contain fewer true contacts as well as some false contacts.\par

\textbf{Relationship between structural quality and inter-chain contact accuracy.} In Appendix \ref{sec:appendix_a}, we further investigate TM-score and $f_{nat}$'s relationship to the precision, recall, and F1 score of \textsc{DRLComplex}'s predicted inter-chain contact inputs. In particular, Figures \ref{figure2}, \ref{figure3}, and \ref{figure4} demonstrate that the quality of \textsc{DRLComplex}'s reconstructed structures largely increases with an increase in the accuracy of its predicted inter-chain contact inputs.\par

\begin{figure}
  \centering
  \includegraphics[width=\linewidth]{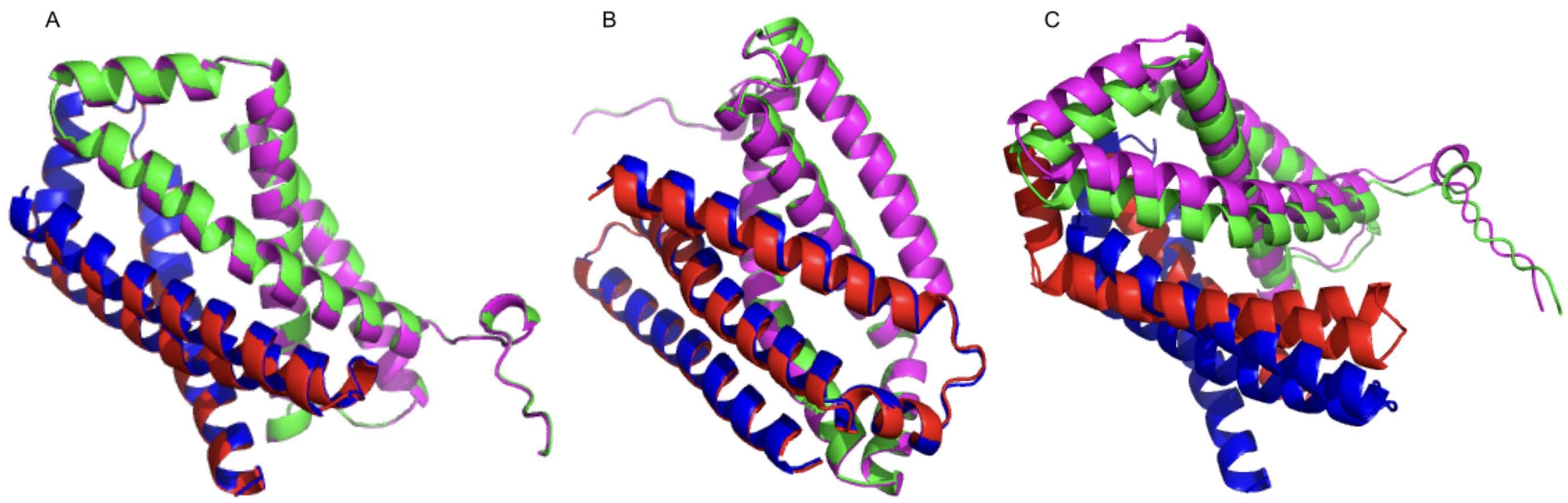}
  \caption{The superimposition of the native structure of a heterodimer (PDB Code: 2WDQ) and the models reconstructed by \textsc{DRLComplex} (i.e., green and red: true structure of the two chains in the dimer, blue and magenta: predicted structure of the two chains): (A) using true tertiary structures and true inter-chain contacts as inputs, (B) using true tertiary structures and predicted inter-chain contacts as input, (C) using predicted tertiary structures and predicted inter-chain contacts as input. TM-score, RMSD, $f_{nat}$, I\_RMSD, L\_RMSD of the model predicted by \textsc{DRLComplex} for (A) are 0.97, 0.94 {\AA}, 97.1 \%, 0.95 {\AA}, and 1.85 {\AA} respectively, for (B) are 0.99, 0.65 {\AA}, 92.1 \%, 0.58 {\AA}, and 1.12 {\AA}, respectively, and for (C) are 0.75, 6.03 {\AA}, 1 \%, 6.07 {\AA}, and 10.6 {\AA}, respectively.}
  \label{figure1}
\end{figure}

\textbf{Visualization.} Figure \ref{figure1} illustrates the reconstructed quaternary structure for a heterodimer alongside its corresponding true structure.\par

\textbf{Limitations.} Currently, \textsc{DRLComplex} adopts a self-play mechanism to speed up its predictions. However, in future works, we would like to investigate alternative data-efficient means of decreasing the time required to train \textsc{DRLComplex} models.\par

\section{Conclusion}
We developed a novel deep reinforcement learning method (\textsc{DRLComplex}) to reconstruct the quaternary structures from inter-chain contacts. It is the first such method to use the reinforcement self-learning to adjust the position of one protein (a ligand) with respect to another (a receptor), considering both short and long-term rewards to search for near-native quaternary structures. Our experiments demonstrate that \textsc{DRLComplex} can successfully generate high-quality structural models for nearly all dimers tested when true inter-chain contacts are provided. We have shown that \textsc{DRLComplex} can also construct the structural model of protein complexes from predicted contacts with reasonable quality if such contact inputs are of sufficient accuracy. Evaluating \textsc{DRLComplex} under increasingly-difficult real-world scenarios demonstrates that our method achieves state-of-the-art results for reconstruction of protein dimer structures using inter-chain contacts as input. In future works, we plan to pre-train \textsc{DRLComplex} on a large set of representative protein dimers to enable the model to directly predict the assembly actions for two units of any new dimer into near-native structures without using on-the-fly self-learning to speed up the model's predictions. As such, we believe \textsc{DRLComplex} will help encourage the adoption of novel deep learning algorithms in drug discovery.

\section{Broader Impacts}
Built upon a simple architecture, our novel method, \textsc{DRLComplex}, is a good case to prove that a self-learning algorithm can learn to reconstruct protein complexes without any guidance or domain knowledge, as accurately as possible. This method can significantly enhance the ability to reconstruct the protein complexes from disoriented or disordered protein clusters by rearranging them in their native conformation state. As explained in our future work, further training of \textsc{DRLComplex} on larger datasets and extending it to predict structures of higher-order multimers could significantly increase the accuracy of its predictions with drastic improvements in speed in comparison to the other simulation methods such as Monte Carlo and simulated annealing methods. Our ideas may also inspire other researchers to conduct new research into the applications of deep reinforcement learning to similar fields like gene regulation, protein-ligand interaction, and drug design, with the introduction of better and more complex architectures as suited to their field. To enable reproducibility and extensions of our work, we have made our code open-source.

\bibliographystyle{unsrtnat}
\bibliography{references}

\appendix

\newpage
\section{Additional Results}
\label{sec:appendix_a}

\begin{figure}[H]
  \centering
  \includegraphics[width=1\linewidth]{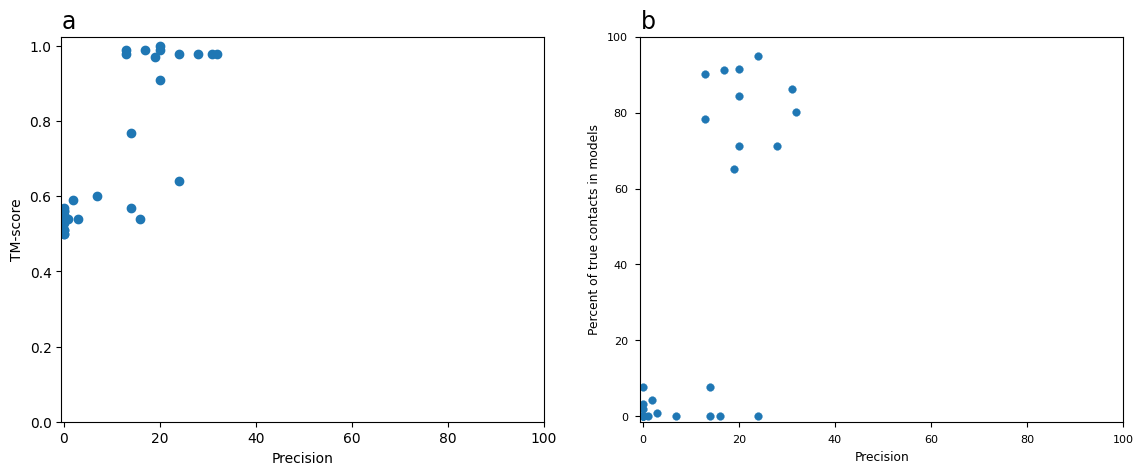}
  \caption{TM-score and $f_{nat}$ of the models generated by \textsc{DRLComplex} VS the precision of the predicted inter-chain contacts on the CASP\_CAPRI homodimer dataset. The true tertiary structures of monomers are used as input.}
  \label{figure2}
\end{figure}

\begin{figure}[H]
  \centering
  \includegraphics[width=1\linewidth]{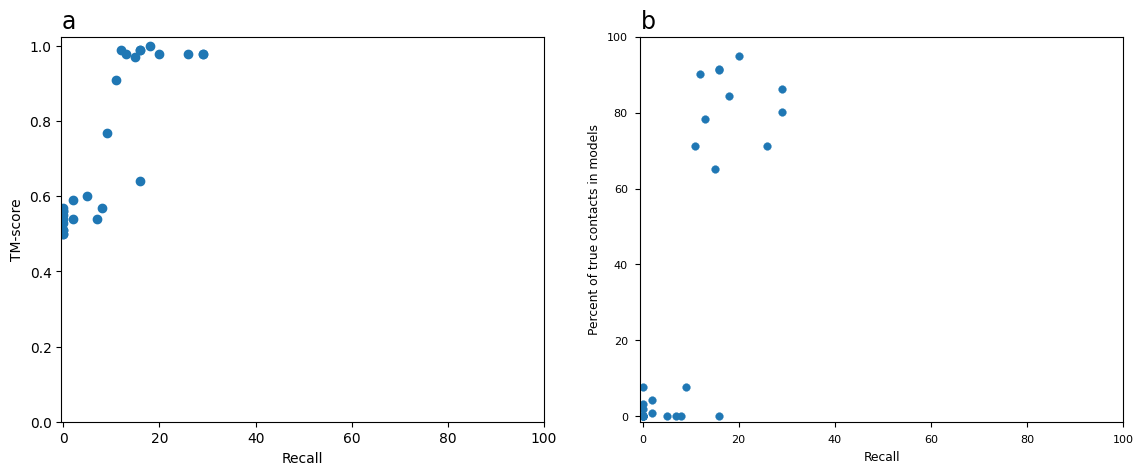}
  \caption{TM-score and $f_{nat}$ of the models generated by \textsc{DRLComplex} VS the recall of the predicted inter-chain contacts on the CASP\_CAPRI homodimer dataset. The true tertiary structures of monomers are used as input.}
  \label{figure3}
\end{figure}

\begin{figure}[H]
  \centering
  \includegraphics[width=1\linewidth]{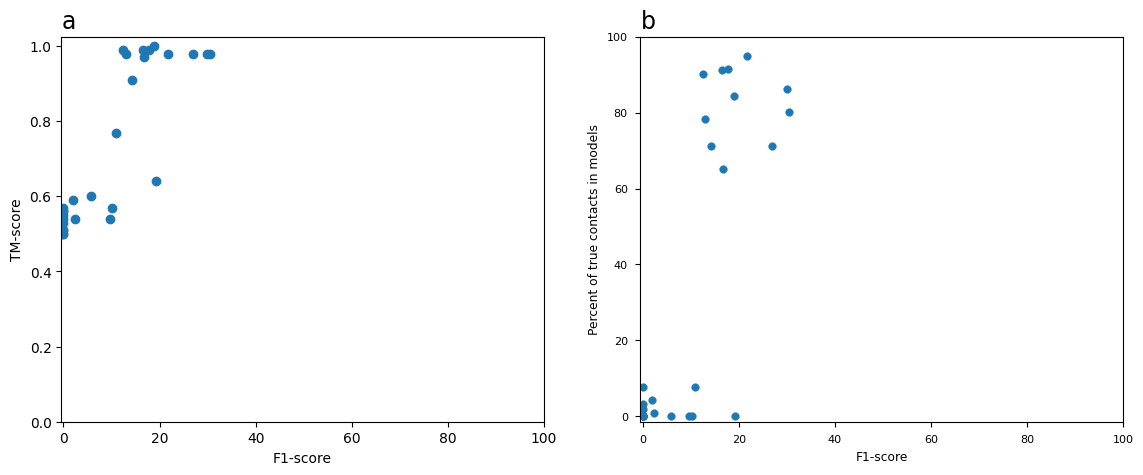}
  \caption{TM-score and $f_{nat}$ of the models generated by \textsc{DRLComplex} VS the F1 score of the predicted inter-chain contacts on the CASP\_CAPRI homodimer dataset. The true tertiary structures of monomers are used as input.}
  \label{figure4}
\end{figure}

\textbf{Inter-chain contact importance study.} In Figures \ref{figure2}, \ref{figure3}, and \ref{figure4}, we inspect the relationship between TM-score (and $f_{nat}$) and the precision, recall, and F1 score of \textsc{DRLComplex}'s predicted inter-chain contact inputs, to better understand the factors influencing \textsc{DRLComplex}'s success in reconstructing quaternary structures. In short, we find that as inter-chain contact accuracy increases, so does the structural quality of \textsc{DRLComplex}'s quaternary reconstructions.\par

\newpage
\textbf{Expanded structure quality results.} In this section, we describe detailed evaluations done on each target of both the CASP\_CAPRI homodimer and Std\_32 heterodimer datasets. The performance of \textsc{DRLComplex} on all the targets of the CASP\_CAPRI homodimer dataset is reported in Tables \ref{stable1}, \ref{stable2}, and \ref{stable3}. Similarly, the performance of \textsc{DRLComplex} on all the targets of the Std\_32 heterodimer dataset is reported in Tables \ref{stable4}, \ref{stable5}, and \ref{stable6}.\par

\begin{table}[H]
  \caption{The RMSD, TM-score, f$_{nat}$, I\_RMSD, and L\_RMSD of $\textsc{DRLComplex}$ for individual targets in the CASP-CAPRI dataset using true contacts and true tertiary structures as inputs. From the table, it can be seen that the RMSD values range from 0.12 to 3.57, with a mean of 0.375 and a median of 0.235. The TM-score values range from 0.768 to 0.999, with an average of 0.989. For the f$_{nat}$ metric, values range from 0.96 to 1 with an average of 0.99. I\_RMSD has a minimum value of 0.125, a maximum value of 2.979, an average value of 0.326, and a median value of 0.241. Lastly, the L\_RMSD metric has values ranging from 0.261 to 7.106, with an average of 0.8 and a median score of 0.533.}
  \label{stable1}
  \centering
  \begin{tabular}{llllll}
    \toprule
    \multicolumn{6}{c}{Optimal Scenario}\\
    \cmidrule(r){2-6}
    \textbf{Targets}  & \textbf{RMSD} ( $\downarrow$ ) & \textbf{TM-score} ( $\uparrow$ ) & \textbf{f$_{nat}$} (\%, $\uparrow$ ) & \textbf{I\_RMSD} ( $\downarrow$ ) & \textbf{L\_RMSD} ( $\downarrow$ ) \\
    \midrule
        T0759   & 0.22   & 0.998    & 100         & 0.224   & 0.527    \\
        T0764   & 0.18   & 0.9995   & 100         & 0.193   & 0.391    \\
        T0770   & 0.26   & 0.9992   & 98.4        & 0.262   & 0.562    \\
        T0776   & 0.19   & 0.9992   & 100         & 0.2     & 0.475    \\
        T0780   & 0.15   & 0.9995   & 97.8        & 0.157   & 0.324    \\
        T0792   & 0.56   & 0.9851   & 97.7        & 0.434   & 1.393    \\
        T0801   & 0.22   & 0.9993   & 99.3        & 0.256   & 0.535    \\
        T0805   & 0.16   & 0.9994   & 97.4        & 0.167   & 0.338    \\
        T0811   & 0.23   & 0.9991   & 98.5        & 0.23    & 0.516    \\
        T0813   & 0.12   & 0.9998   & 98.6        & 0.125   & 0.261    \\
        T0815   & 3.57   & 0.768    & 100         & 2.979   & 7.106    \\
        T0819   & 0.2    & 0.9994   & 99.2        & 0.209   & 0.424    \\
        T0825   & 0.37   & 0.9967   & 100         & 0.326   & 0.922    \\
        T0843   & 0.24   & 0.9993   & 99.4        & 0.252   & 0.532    \\
        T0847   & 0.21   & 0.9989   & 100         & 0.218   & 0.468    \\
        T0849   & 0.25   & 0.9988   & 99.2        & 0.264   & 0.537    \\
        T0851   & 0.22   & 0.9995   & 99          & 0.222   & 0.45     \\
        T0852   & 0.26   & 0.9991   & 98.6        & 0.279   & 0.636    \\
        T0893   & 0.38   & 0.9852   & 99          & 0.278   & 0.89     \\
        T0965   & 0.38   & 0.998    & 100         & 0.35    & 0.804    \\
        T0966   & 0.24   & 0.9994   & 100         & 0.272   & 0.633    \\
        T0976   & 0.23   & 0.9991   & 100         & 0.209   & 0.492    \\
        T0984   & 0.29   & 0.9993   & 99.1        & 0.227   & 0.663    \\
        T0999D1 & 0.38   & 0.9983   & 98.1        & 0.37    & 0.863    \\
        T0999D4 & 0.28   & 0.9986   & 100         & 0.265   & 0.749    \\
        T1003   & 0.18   & 0.9996   & 98.8        & 0.189   & 0.393    \\
        T1006   & 0.34   & 0.9941   & 96.8        & 0.347   & 0.742    \\
        T1032   & 0.2    & 0.999    & 98.2        & 0.208   & 0.433    \\
    \midrule
        Mean    & 0.3753 & 0.9895   & 99.05       & 0.2197  & 0.8235  \\
    \bottomrule
  \end{tabular}
\end{table}

\begin{table}[H]
  \caption{Detailed results (RMSD, TM-score, f$_{nat}$, I\_RMSD, and L\_RMSD) of $\textsc{DRLComplex}$ for the CASP-CAPRI dataset using predicted inter-chain contacts and true tertiary structure. The TM-score values range from 0.50 to 0.99, with an average value of 0.73.}
  \label{stable2}
  \centering
  \begin{tabular}{llllll}
    \toprule
    \multicolumn{6}{c}{Suboptimal Scenario}\\
    \cmidrule(r){2-6}
    \textbf{Targets}  & \textbf{RMSD} ( $\downarrow$ ) & \textbf{TM-score} ( $\uparrow$ ) & \textbf{f$_{nat}$} (\%, $\uparrow$ ) & \textbf{I\_RMSD} ( $\downarrow$ ) & \textbf{L\_RMSD} ( $\downarrow$ ) \\
    \midrule
        T0976   & 18.69   & 0.54     & 0           & 17.8    & 36       \\
        T0776   & 4.65    & 0.77     & 7.6         & 5.706   & 13.31    \\
        T0813   & 0.98    & 0.99     & 90.1        & 0.925   & 2.366    \\
        T0852   & 29.34   & 0.57     & 0           & 21.71   & 48.98    \\
        T0966   & 27.26   & 0.53     & 0           & 12.55   & 62.44    \\
        T1003   & 0.89    & 0.99     & 91.3        & 0.833   & 1.719    \\
        T0819   & 0.62    & 1        & 84.3        & 0.631   & 1.211    \\
        T0965   & 16.28   & 0.59     & 4.3         & 13.86   & 28.57    \\
        T0792   & 14.57   & 0.5      & 0           & 14.93   & 40.23    \\
        T0851   & 1.3     & 0.98     & 80.2        & 0.858   & 1.819    \\
        T0815   & 10.52   & 0.51     & 0           & 11.9    & 32.47    \\
        T0770   & 24.99   & 0.54     & 0           & 23.16   & 50.39    \\
        T1032   & 18.13   & 0.54     & 0.9         & 17.07   & 29.24    \\
        T0999D1 & 14.42   & 0.64     & 0           & 12.66   & 27.34    \\
        T0805   & 1.07    & 0.98     & 78.4        & 1.08    & 2.326    \\
        T0780   & 22.7    & 0.55     & 0           & 21.52   & 55.84    \\
        T1006   & 16.05   & 0.5      & 0           & 19.34   & 50.87    \\
        T0843   & 0.92    & 0.99     & 91.4        & 0.942   & 1.466    \\
        T0893   & 0.85    & 0.98     & 94.8        & 0.92    & 1.94     \\
        T0811   & 1.08    & 0.98     & 86.3        & 0.901   & 3.675    \\
        T0984   & 39.94   & 0.56     & 1.8         & 31.43   & 72.91    \\
        T0849   & 1.07    & 0.98     & 71.2        & 1.086   & 3.437    \\
        T0764   & 13.7    & 0.56     & 3.2         & 12.97   & 33.29    \\
        T0759   & 14.09   & 0.51     & 0           & 12.38   & 38.39    \\
        T0999D4 & 2.48    & 0.91     & 71.1        & 1.661   & 10.3     \\
        T0825   & 17.39   & 0.57     & 7.8         & 14.91   & 32.18    \\
        T0847   & 17.08   & 0.6      & 0           & 16.69   & 55.77    \\
        T0801   & 1.67    & 0.97     & 65.24       & 1.72    & 3.98     \\
    \midrule
        Mean    & 11.8832 & 0.73     & 33.2121     & 10.4336 & 26.5163 \\
    \bottomrule
  \end{tabular}
\end{table}

\begin{table}[H]
  \caption{Detailed results (RMSD, TM-score, f$_{nat}$, I\_RMSD, L\_RMSD, and monomer TM-score) of $\textsc{DRLComplex}$ for the CASP-CAPRI dataset using predicted inter-chain contacts and predicted tertiary structures. The TM-score values range from 0.36 to 0.97, with an average value of 0.64. The average TM-score of the monomers predicted by AlphaFold 2 is 0.95.}
  \label{stable3}
  \centering
  \begin{tabular}{lllllll}
    \toprule
    \multicolumn{7}{c}{Realistic Scenario}\\
    \cmidrule(r){2-7}
    \textbf{Targets}  & \textbf{RMSD} ( $\downarrow$ ) & \textbf{TM-score} ( $\uparrow$ ) & \textbf{f$_{nat}$} (\%, $\uparrow$ ) & \textbf{I\_RMSD} ( $\downarrow$ ) & \textbf{L\_RMSD} ( $\downarrow$ ) & TM-score (monomer) \\
    \midrule
        T1003   & 0.58  & 0.92     & 78  & 0.51    & 0.8     & 0.9964   \\
    T0984   & 33.57 & 0.48     & 2  & 27.17   & 57.5    & 0.9805  \\
    T0792   & 12.7  & 0.49     & 7 & 12.16   & 30.47   & 0.9585   \\
    T0805   & 1.83  & 0.94     & 75 & 1.85    & 2.39    & 0.9675 \\
    T0851   & 2.03  & 0.93     & 73 & 1.68    & 3.01    & 0.9687 \\
    T0999D1 & 16.52 & 0.51     & 2   & 13.24   & 34.22   & 0.9708 \\
    T0815   & 14.7  & 0.49     & 1 & 11.29   & 47.88   & 0.9789 \\
    T0759   & 15.62 & 0.4      & 7 & 12.42   & 35.6    & 0.7475  \\
    T0893   & 20.09 & 0.36     & 32   & 15.39   & 54.24   & 0.7178 \\
    T0825   & 18.23 & 0.49     & 0   & 15.64   & 30.46   & 0.9655   \\
    T0819   & 1.94  & 0.88     & 47   & 2.28    & 2.22    & 0.9705   \\
    T1006   & 11.39 & 0.47     & 0    & 11.93   & 33.96   & 0.9859 \\
    T0966   & 32.4  & 0.48     & 9  & 30.28   & 76.42   & 0.9529 \\
    T0770   & 11.7  & 0.53     & 8    & 13.61   & 28.94   & 0.9777 \\
    T0843   & 1.04  & 0.95     & 63  & 0.9     & 1.72    & 0.988 \\
    T0999D4 & 7.03  & 0.59     & 0    & 8.09    & 17.99   & 0.9767 \\
    T0780   & 22.17 & 0.46     & 4  & 21.1    & 56.72   & 0.9599 \\
    T0976   & 18.16 & 0.52     & 3   & 16.99   & 34.17   & 0.9777  \\
    T0811   & 0.88  & 0.96     & 77   & 0.87    & 2.24    & 0.993 \\
    T0852   & 33.26 & 0.45     & 6    & 21.91   & 52.15   & 0.9334 \\
    T1032   & 6.25  & 0.6      & 45  & 5.6     & 8.49    & 0.7 \\
    T0776   & 4.8   & 0.75     & 26   & 3.57    & 18.46   & 0.9765 \\
    T0801   & 2.12  & 0.88     & 57   & 2.3     & 4.11    & 0.9648  \\
    T0813   & 1.14  & 0.97     & 69  & 1.34    & 1.2     & 0.9818 \\
    T0764   & 15.14 & 0.55     & 1  & 16.74   & 27.8    & 0.9882  \\
    T0849   & 1.38  & 0.88     & 64  & 1.14    & 1.79    & 0.9727  \\
    T0965   & 15.64 & 0.57     & 1  & 13.24   & 27.57   & 0.9851  \\
    T0847   & 18.81 & 0.52     & 2  & 17.41   & 50.72   & 0.9901  \\
    \midrule
    Mean    & 12.18 & 0.64     & 27.1   & 10.74   & 26.54   & 0.95 \\              
    \bottomrule
  \end{tabular}
\end{table}

\begin{table}[H]
  \caption{Detailed results (RMSD, TM-score, f$_{nat}$, I\_RMSD, and L\_RMSD) of $\textsc{DRLComplex}$ on the Std\_32 dataset with true inter-chain contacts and true tertiary structures as inputs. The average TM-score is 0.987, with a minimum value of 0.97 and a maximum value of 0.99.}
  \label{stable4}
  \centering
  \begin{tabular}{llllll}
    \toprule
    \multicolumn{6}{c}{Optimal Scenario}\\
    \cmidrule(r){2-6}
    \textbf{Targets}  & \textbf{RMSD} ( $\downarrow$ ) & \textbf{TM-score} ( $\uparrow$ ) & \textbf{f$_{nat}$} (\%, $\uparrow$ ) & \textbf{I\_RMSD} ( $\downarrow$ ) & \textbf{L\_RMSD} ( $\downarrow$ ) \\
    \midrule
        3RRLA\_3RRLB & 0.93 & 0.98     & 98.2 & 0.902   & 2.441    \\
        2NU9A\_2NU9B & 0.95 & 0.99     & 96.1 & 0.995   & 1.678    \\
        1EP3A\_1EP3B & 0.84 & 0.99     & 99.3  & 1.044   & 1.541    \\
        2Y69B\_2Y69C & 0.74 & 0.99     & 84.1  & 0.756   & 2.28     \\
        3RPFA\_3RPFC & 0.93 & 0.97     & 50    & 0.864   & 2.475    \\
        1TYGB\_1TYGA & 0.81 & 0.99     & 99.2  & 0.927   & 1.986    \\
        3MMLA\_3MMLB & 0.96 & 0.99     & 94.8  & 0.932   & 2.063    \\
        2VPZA\_2VPZB & 0.91 & 0.99     & 93.8   & 1.172   & 2.305    \\
        2Y69A\_2Y69C & 0.86 & 0.99     & 82.6  & 0.916   & 2.123    \\
        1I1QA\_1I1QB & 0.87 & 0.99     & 98.8   & 0.906   & 2.069    \\
        2Y69A\_2Y69B & 0.75 & 0.99     & 86.3    & 0.842   & 1.96     \\
        1EFPA\_1EFPB & 0.97 & 0.99     & 97.5  & 1.04    & 3.152    \\
        1W85A\_1W85B & 0.77 & 0.99     & 99.3   & 0.76    & 2.263    \\
        1ZUNA\_1ZUNB & 0.93 & 0.99     & 98.11  & 0.954   & 1.761    \\
        3PNLA\_3PNLB & 0.87 & 0.99     & 99.4  & 0.782   & 2.077    \\
        3OAAH\_3OAAG & 0.9  & 0.99     & 93.8   & 1.008   & 1.851    \\
        3G5OA\_3G5OB & 0.86 & 0.97     & 90.2    & 0.839   & 1.685    \\
        2WDQC\_2WDQD & 0.94 & 0.97     & 97.1   & 0.95    & 1.853    \\
        1BXRA\_1BXRB & 0.96 & 0.99     & 97.6   & 1.14    & 3.119    \\
        1RM6A\_1RM6B & 0.79 & 0.99     & 96.5   & 0.604   & 1.892    \\
        1QOPA\_1QOPB & 0.78 & 0.99     & 97.2   & 0.761   & 1.625    \\
        1B70A\_1B70B & 0.96 & 0.99     & 92.4   & 1.123   & 2.165    \\
        3A0RA\_3A0RB & 0.8  & 0.99     & 98.1  & 0.699   & 2.017    \\
        2ONKA\_2ONKC & 0.9  & 0.99     & 96.4   & 0.535   & 1.932    \\
        2D1PB\_2D1PC & 0.92 & 0.97     & 83.7  & 0.935   & 2.325    \\
        4HR7A\_4HR7B & 0.93 & 0.99     & 82.6   & 1.03    & 2.472    \\
        1RM6A\_1RM6C & 0.95 & 0.99     & 74.5    & 1.323   & 2.434    \\
        3IP4B\_3IP4C & 0.93 & 0.99     & 79.7   & 0.946   & 2.628    \\
        3IP4A\_3IP4C & 0.97 & 0.99     & 74.3  & 1.158   & 2.77     \\
        1RM6B\_1RM6C & 0.81 & 0.99     & 69.3    & 0.9     & 1.696    \\
        \midrule
        Mean         & 0.88 & 0.987    & 90.03   & 0.92    & 2.15    \\
    \bottomrule
  \end{tabular}
\end{table}

\begin{table}[H]
  \caption{Detailed results (RMSD, TM-score, f$_{nat}$, I\_RMSD, and L\_RMSD) of $\textsc{DRLComplex}$ on our 31 Std\_32 heterodimers using predicted inter-chain contacts as inputs. We note that true monomer structures are used in this experiment. Also, targets that do not contain any inter-chain contacts are discarded.}
  \label{stable5}
  \centering
  \begin{tabular}{llllll}
    \toprule
    \multicolumn{6}{c}{Suboptimal Scenario}\\
    \cmidrule(r){2-6}
    \textbf{Targets}  & \textbf{RMSD} ( $\downarrow$ ) & \textbf{TM-score} ( $\uparrow$ ) & \textbf{f$_{nat}$} (\%, $\uparrow$ ) & \textbf{I\_RMSD} ( $\downarrow$ ) & \textbf{L\_RMSD} ( $\downarrow$ ) \\
    \midrule
        1EFPA\_1EFPB & 3.19    & 0.87     & 15 & 3.24    & 7.82     \\
        1EP3A\_1EP3B & 8.08    & 0.67     & 11 & 6.45    & 17.9     \\
        1I1QA\_1I1QB & 19.29   & 0.76     & 5 & 21.2    & 55.52    \\
        1QOPA\_1QOPB & 20.38   & 0.62     & 5 & 23.72   & 51.28    \\
        1W85A\_1W85B & 0.73    & 1        & 80 & 0.48    & 2.02     \\
        1ZUNA\_1ZUNB & 15.48   & 0.65     & 8 & 16.54   & 30.22    \\
        2D1PB\_2D1PC & 14.89   & 0.49     & 4  & 14.17   & 34.35    \\
        2NU9A\_2NU9B & 2.63    & 0.89     & 48 & 2.23    & 5.17     \\
        2ONKA\_2ONKC & 27.47   & 0.52     & 2 & 25.26   & 65.44    \\
        2VPZA\_2VPZB & 34.39   & 0.84     & 2 & 32.21   & 92.48    \\
        2WDQC\_2WDQD & 0.65    & 1        & 92 & 0.58    & 1.12     \\
        2Y69A\_2Y69B & 26.1    & 0.68     & 8 & 27.41   & 65.18    \\
        2Y69A\_2Y69C & 25.66   & 0.61     & 6 & 25.21   & 59.85    \\
        2Y69B\_2Y69C & 20.94   & 0.53     & 10 & 16.56   & 65.58    \\
        3A0RA\_3A0RB & 17.56   & 0.82     & 7 & 20.38   & 43.67    \\
        3G5OA\_3G5OB & 8.85    & 0.66     & 4 & 9.41    & 13.83    \\
        3IP4A\_3IP4B & 25.71   & 0.5      & 8 & 20.64   & 57.94    \\
        3IP4A\_3IP4C & 12.22   & 0.87     & 7 & 14.91   & 31.5     \\
        3IP4B\_3IP4C & 0.67    & 1        & 89 & 0.44    & 2.16     \\
        3MMLA\_3MMLB & 2.64    & 0.9      & 27 & 2.36    & 5.62     \\
        3OAAH\_3OAAG & 3.79    & 0.93     & 35 & 4.22    & 7.93     \\
        3PNLA\_3PNLB & 6.95    & 0.7      & 6  & 6.29    & 17.11    \\
        3RPFA\_3RPFC & 17.47   & 0.73     & 7  & 9.07    & 53.93    \\
        3RRLA\_3RRLB & 12.91   & 0.62     & 5 & 12.63   & 25.17    \\
        4HR7A\_4HR7B & 8.35    & 0.95     & 12 & 9.6     & 23.34    \\
        1B70A\_1B70B & 16.77   & 0.82     & 2 & 20.58   & 34.72    \\
        1BXRA\_1BXRB & 20.79   & 0.81     & 6  & 20.82   & 49.85    \\
        1RM6A\_1RM6B & 26.13   & 0.74     & 6   & 25.87   & 59.09    \\
        1RM6A\_1RM6C & 17.4    & 0.74     & 9  & 18.68   & 48.44    \\
        1RM6B\_1RM6C & 13.13   & 0.65     & 4  & 12.86   & 29.88    \\
        1TYGB\_1TYGA & 0.53    & 0.91     & 79  & 0.34    & 0.87     \\
        \midrule
        Mean         & 13.9274 & 0.7574   & 19.6451    & 13.689  & 34.1606  \\
    \bottomrule
  \end{tabular}
\end{table}

\begin{table}[H]
  \caption{Detailed results (RMSD, TM-score, f$_{nat}$, I\_RMSD, L\_RMSD, TM-score of the ligand, and TM-score of the receptor) of $\textsc{DRLComplex}$ on the Std\_32 dataset with predicted inter-chain contacts and predicted tertiary structures as inputs. The average TM-score is 0.74, with a minimum value of 0.5 and a maximum value of 1.}
  \label{stable6}
  \centering
  \begin{tabular}{llllllll}
    \toprule
    \multicolumn{8}{c}{Realistic Scenario}\\
    \cmidrule(r){2-8}
    \textbf{Targets}  & \textbf{RMSD}  & \textbf{TM-score}  & \textbf{f$_{nat}$}  & \textbf{I\_RMSD} & \textbf{L\_RMSD} & \textbf{TM-score} & \textbf{TM-score}  \\
     & ( $\downarrow$ ) & ( $\uparrow$ ) & (\%, $\uparrow$ ) &  ( $\downarrow$ ) &  ( $\downarrow$ )  &  \textbf{(Ligand)} ( $\uparrow$ )  & \textbf{(Receptor)} ( $\uparrow$ )\\
    \midrule
        1EFPA\_1EFPB & 2.83   & 0.89     & 25     & 3.21    & 6.08    & 0.98  & 0.96 \\
        1EP3A\_1EP3B & 5.07   & 0.86     & 16     & 4.48    & 11.12   & 0.97   & 0.98 \\
        1I1QA\_1I1QB & 20.79  & 0.67     & 5      & 21.52   & 62.06   & 0.97   & 0.97  \\
        1QOPA\_1QOPB & 20.22  & 0.64     & 7      & 23.35   & 50.21   & 0.97   & 0.99   \\
        1W85A\_1W85B & 4.64   & 0.82     & 11     & 3.33    & 5.88    & 0.99   & 0.99  \\
        1ZUNA\_1ZUNB & 16.22  & 0.67     & 3      & 17.45   & 27.81   & 0.93   & 0.95   \\
        2D1PB\_2D1PC & 10.61  & 0.61     & 0      & 10.77   & 28.47   & 0.99   & 0.99  \\
        2NU9A\_2NU9B & 1.92   & 1        & 52     & 1.83    & 3.41    & 0.99  & 0.97   \\
        2VPZA\_2VPZB & 32.16  & 0.81     & 1      & 30.17   & 89.68   & 0.98   & 0.94   \\
        2WDQC\_2WDQD & 6.03   & 0.75     & 1      & 6.07    & 10.6    & 0.96  & 0.98   \\
        2Y69A\_2Y69B & 24.2   & 0.77     & 9      & 23.51   & 58.93   & 0.99   & 0.98  \\
        2Y69A\_2Y69C & 19.97  & 0.79     & 1      & 21.49   & 40.44   & 0.99   & 0.99  \\
        2Y69B\_2Y69C & 18.91  & 0.56     & 3      & 7.11    & 58      & 0.98  & 0.99   \\
        3A0RA\_3A0RB & 21.38  & 0.6      & 0      & 20.33   & 53.29   & 0.77  & 0.9    \\
        3G5OA\_3G5OB & 14.11  & 0.54     & 3      & 14.14   & 26.31   & 0.9    & 0.96    \\
        3IP4A\_3IP4B & 20.31  & 0.51     & 4      & 14.58   & 81.33   & 0.99     & 0.88   \\
        3IP4A\_3IP4C & 11.57  & 0.82     & 1      & 14.83   & 29.61   & 0.99   & 0.92  \\
        3IP4B\_3IP4C & 5.01   & 0.95     & 34     & 2.18    & 6.99    & 0.88    & 0.92  \\
        3MMLA\_3MMLB & 3.83   & 0.81     & 40     & 3.2     & 9.24    & 0.99   & 0.97    \\
        3OAAH\_3OAAG & 14.53  & 0.81     & 26     & 17.55   & 26.62   & 0.59    & 0.97   \\
        3PNLA\_3PNLB & 7.07   & 0.69     & 1      & 6.59    & 18.67   & 0.98    & 0.99  \\
        3RPFA\_3RPFC & 16.36  & 0.64     & 10     & 7.73    & 54.69   & 0.97   & 0.96   \\
        3RRLA\_3RRLB & 7.66   & 0.73     & 9      & 8.03    & 12.29   & 0.99    & 0.93  \\
        4HR7A\_4HR7B & 19.39  & 0.73     & 6      & 20.02   & 57.84   & 0.9     & 0.96   \\
        1B70A\_1B70B & 17.59  & 0.83     & 2      & 21.32   & 36.1    & 0.97   & 0.95    \\
        1BXRA\_1BXRB & 20.93  & 0.79     & 2      & 20.99   & 50.06   & 0.99   & 0.99   \\
        1RM6A\_1RM6B & 26.29  & 0.6      & 4      & 25.94   & 59.11   & 0.99   & 0.99  \\
        1RM6A\_1RM6C & 17.74  & 0.85     & 0      & 18.84   & 49.08   & 0.99   & 0.98    \\
        1RM6B\_1RM6C & 13.9   & 0.63     & 5      & 12.95   & 29.93   & 0.99  & 0.98   \\
        1TYGB\_1TYGA & 0.4    & 0.97     & 81     & 0.41    & 0.82    & 0.9    & 0.96    \\
        2ONKA\_2ONKC & 28.65  & 0.5      & 4      & 25.68   & 66.83   & 0.97  & 0.93   \\
        \midrule
        Mean         & 14.525 & 0.7367   & 11.806 & 13.858  & 36.177  & 0.95 & 0.962   \\  
    \bottomrule
  \end{tabular}
\end{table}

\end{document}